%% file: main.tex
\documentclass{article} % For LaTeX2e
\usepackage{iclr2027_conference,times}
\usepackage{helvet}
\usepackage{courier}
\usepackage{graphicx}
\usepackage{hyperref}
\usepackage{url}
\usepackage{caption}
\usepackage{wrapfig}

\usepackage{algorithm}
\usepackage{algorithmic}
\usepackage{amsmath}
\usepackage{amsfonts}

\usepackage{booktabs}
\usepackage{multirow}

\usepackage{colortbl}
\usepackage{xcolor}

\usepackage{arydshln}

\newcommand{\std}[1]{{\fontsize{6pt}{6pt}\selectfont (#1)}}
\usepackage{pgfplots}
\pgfplotsset{compat=1.18}
\usepgfplotslibrary{fillbetween}
\usetikzlibrary{plotmarks}

\usepackage{newfloat}
\usepackage{listings}
\DeclareCaptionStyle{ruled}{labelfont=normalfont,labelsep=colon,strut=off}
\floatstyle{ruled}
\newfloat{listing}{tb}{lst}{}
\floatname{listing}{Listing}
\title{Learning from Environmental Feedback: Credit  Assignment across Multiple Timescales for Agentic Reinforcement Learning}

\author{
Yifu Huo$^1$,
Shunjie Xing$^1$,
Chenglong Wang$^{1,2}$,
Peinan Feng$^1$,
Qiaozhi He$^1$,
Yan Ding$^1$, \\
\textbf{
Anxiang Ma$^1$,
Yuxin Gao$^1$,
Tongran Liu$^3$,
Tong Xiao$^{1,2}$\thanks{Corresponding author.},
Jingbo Zhu$^{1,2}$} \\
\ $^1$School of Computer Science and Engineering, Northeastern University, Shenyang, China \\
\ $^2$NiuTrans Research, Shenyang, China \\
\ $^3$CAS Key Laboratory of Behavioral Science, Institute of Psychology, CAS, Beijing, China \\
\ \texttt{xiaotong@mail.neu.edu.cn}
}

\iclrfinalcopy

\begin{document}

\maketitle

\begin{abstract}
Agentic reinforcement learning (RL) often suffers from delayed and sparse rewards in real-world environments. 
A promising solution to this challenge is credit assignment, which aims to decompose trajectory-level rewards and provide more fine-grained supervision for intermediate decisions. 
However, existing credit assignment approaches ignore the rich process information naturally generated during environment interaction, e.g., interaction history.
We argue that such information provides valuable supervision for identifying the contribution of individual actions. 
To this end, we propose \textbf{\underline{E}}nvironmental \textbf{\underline{F}}eedback-based \textbf{\underline{C}}redit \textbf{\underline{A}}ssignment (EFCA), a multi-timescale credit assignment approach for long-horizon agentic RL. 
EFCA complements the long-term outcome signal with two environment-grounded process signals: a \textit{short-term feedback signal} that captures the immediate effect of the current action and \textit{a medium-term state-history signal} that identifies ineffective patterns from recent interactions. 
Both signals are directly extracted from environment feedback and integrated through a return reweighting mechanism. 
Experiments on ALFWorld and WebShop demonstrate that EFCA consistently improves both task success and task quality over strong baselines, highlighting the effectiveness of environment-grounded multi-timescale credit assignment for long-horizon agentic RL.
\end{abstract}

\section{Introduction}

Large language models (LLMs) are increasingly optimized as agents through reinforcement learning in interactive environments \cite{yao2022react,Schick2023ToolformerLM,Gur2023ARW,Shridhar2020ALFWorldAT,Yao2022WebShopTS}. 
Unlike conventional reinforcement learning with verifiable rewards (RLVR),  agentic RL produces outcomes through sequences of heterogeneous actions whose contributions are highly uneven rather than evaluating a single answer or self-contained reasoning path using a rule-based verifier \cite{Cobbe2021TrainingVT,Lightman2023LetsVS,Shao2024DeepSeekMathPT}.
This difference makes rewards sparse and delayed in long-horizon environments, creating a mismatch between step-level decision making and episode-level supervision \cite{yao2022react,Feng2025GroupinGroupPO}.
Therefore, accurate step-level credit assignment is therefore essential for training agentic policies under sparse outcome supervision, especially when meaningful rewards are available only after the entire interaction succeeds or fails \cite{Pignatelli2023ASO,Cheng2026BeyondTA}.

Existing methods address this issue mainly through two forms of stepwise supervision.
The first decomposes trajectory-level outcomes into step-level learning signals \cite{Shao2024DeepSeekMathPT,Feng2025GroupinGroupPO,Wang2026StepPOSP,Cheng2026BeyondTA}.
Although these methods provide finer optimization granularity than trajectory-level training, their credit signals are still inferred indirectly from final outcomes and may therefore struggle to determine which intermediate actions are locally effective. 
The second form introduces denser supervision through parameterized state-value or reward models, as commonly used in policy optimization methods such as PPO and VAPO \cite{Schulman2017ProximalPO,Cobbe2021TrainingVT,Lightman2023LetsVS,Yue2025VAPOEA}.
While such learned critics can provide more informative step-level estimates than terminal rewards alone, they are not freely available: the corresponding value or reward model typically needs to be trained from scratch for each target environment. 
Recent work has explored more generalizable reward modeling paradigms, including generative foundation reward models and scalable reinforcement learning strategies for generative multimodal reward modeling, but these approaches still rely on separately learned reward estimators rather than directly exploiting environment-grounded process feedback \cite{Wang2025GRAMAG,Wang2026MSRLSG}.
Moreover, the overall optimization quality is bounded by the reliability of the critic or reward model, which can introduce a mismatch between critic estimates and the true environment dynamics \cite{Yuan2025WhatsBP,Wang2025ProbingPR}.

These limitations motivate an alternative source of supervision: the environmental signals that are naturally produced during interaction. 
At each step of a trajectory, the environment often returns rich process evidence, including textual feedback, action validity, observations, and state changes \cite{yao2022react,Shinn2023ReflexionLA,Shridhar2020ALFWorldAT,Yao2022WebShopTS}.
Importantly, such signals are directly grounded in the environment and require neither external annotation nor an additional learned evaluator.
Therefore, environmental feedback provides a natural and underexplored basis for constructing step-level credit signals that remain aligned with the actual task dynamics.

In this work, we propose Environmental Feedback-based Credit Assignment (EFCA), a credit assignment method for refining step-level returns with environment-grounded multi-timescale signals.
EFCA complements the long-term outcome signal with two environment-grounded process signals.
The first is a \emph{short-term feedback signal}, which captures whether the current action produces local progress according to the immediate environment response.
The second is a \emph{medium-term state-history signal}, which uses recent action-feedback history to detect repeated ineffective behaviors and short-horizon behavioral loops.
EFCA instantiates these signals as feedback credit and state-history credit, and integrates them through a return reweighting mechanism that modulates the original step-level return used by stepwise policy optimization methods such as GiGPO and HGPO \cite{He2026HierarchyofGroupsPO,Feng2025GroupinGroupPO}.
With this streamlined design, EFCA does not require auxiliary value networks, yet can inject richer process-level supervision into rollout trajectories, thereby mitigating the sparsity and delay of rewards in long-horizon agentic tasks.

We evaluate EFCA on two representative long-horizon agentic benchmarks, ALFWorld and WebShop  \cite{Shridhar2020ALFWorldAT,Yao2022WebShopTS}.
Across both environments and model scales, EFCA achieves competitive performance compared with strong stepwise baselines.
These results demonstrate that complementing long-term outcome supervision with environment-grounded multi-timescale process signals can substantially improve credit assignment for long-horizon agentic reinforcement learning.

\section{Preliminary}

% \begin{figure*}[t]
%   \centering
%   \includegraphics[width=\textwidth,page=1]{figures/return_propagation.pdf.pdf}
%   \caption{Return Propagation in Conventional Policy Optimization.}
%   \label{fig:return-propagation}
% \end{figure*}

\subsection{Stepwise Policy Optimization}
Conventional multi-turn policy optimization for interactive agents generally treats an entire rollout as a monolithic trajectory optimized against a sparse terminal reward \cite{sutton1998reinforcement,Lattimer2024SparseRC,Wei2025ReinforcingMR}. 
Despite its conceptual simplicity, this trajectory-wise paradigm degrades in efficiency as the interaction horizon expands \cite{Kim2026OnTL}. 
Specifically, it incurs severe computational overhead by perpetually carrying forward the full historical context, while the learning signals remain fundamentally constrained by the coarse granularity of the final outcome \cite{Chen2025IterResearchRL}.

Stepwise policy optimization addresses this limitation by shifting the optimization unit from monolithic trajectories to individual steps, augmented by a memory module to retain contextual information \cite{Wang2026StepPOSP,Wang2025SPARLRL}. 
Rather than optimizing only at the rollout level, stepwise methods decompose trajectory-level outcome signals into step-level training targets, so that intermediate actions can be updated at a finer granularity even when supervision is still primarily derived from final outcomes.
This paradigm enhances scalability for long-horizon tasks and enables policy updates at a much finer granularity. 
Existing approaches typically integrate trajectory-level outcomes with step-level grouping or relative comparisons, allowing intermediate actions to be evaluated within similar local contexts  \cite{Feng2025GroupinGroupPO,Cheng2026BeyondTA}. 
In this paper, rather than proposing a new grouping mechanism, we focus on directly enhancing the quality of the step-level supervision signals fed into these optimization pipelines.

\subsection{Credit Assignment in Long-horizon Agentic Tasks}
The main difficulty in long-horizon agentic reinforcement learning is that the final task outcome is a coarse and delayed supervision signal, whereas the agent's behavior is determined by many intermediate actions with highly unequal contributions  \cite{Pignatelli2023ASO,Tan2026HindsightCA}. 
If the final reward is simply propagated back to all steps, two failure modes naturally arise. 
First, low-contribution or redundant actions appearing in successful trajectories may be mistakenly reinforced. 
Second, locally useful actions appearing in failed trajectories may be incorrectly suppressed. 
These problems become even more severe in critic-free settings, where step-level supervision must be constructed without an explicit value network \cite{Feng2025GroupinGroupPO,Cheng2026BeyondTA}.

From the perspective of stepwise optimization, the credit assignment problem can therefore be stated as follows: given an original step-level return, how can we refine it so that it better reflects the local usefulness and historical efficiency of each action? 
Although recent studies have improved the efficiency of reinforcement learning for sequence generation, the central difficulty in long-horizon agentic settings remains the construction of accurate step-level supervision under sparse and delayed environmental rewards \cite{Wang2023ESRLES,Huo2026SPSSP}.
Our work addresses this question by introducing a customized credit signal that complements the original return with two specific forms of step-level evidence: immediate environmental feedback and repetition-aware historical failure patterns. 
This preliminary perspective directly motivates the method introduced in the next section.

\section{Method}

\begin{figure*}[t]
  \centering
  \includegraphics[width=\textwidth,page=1]{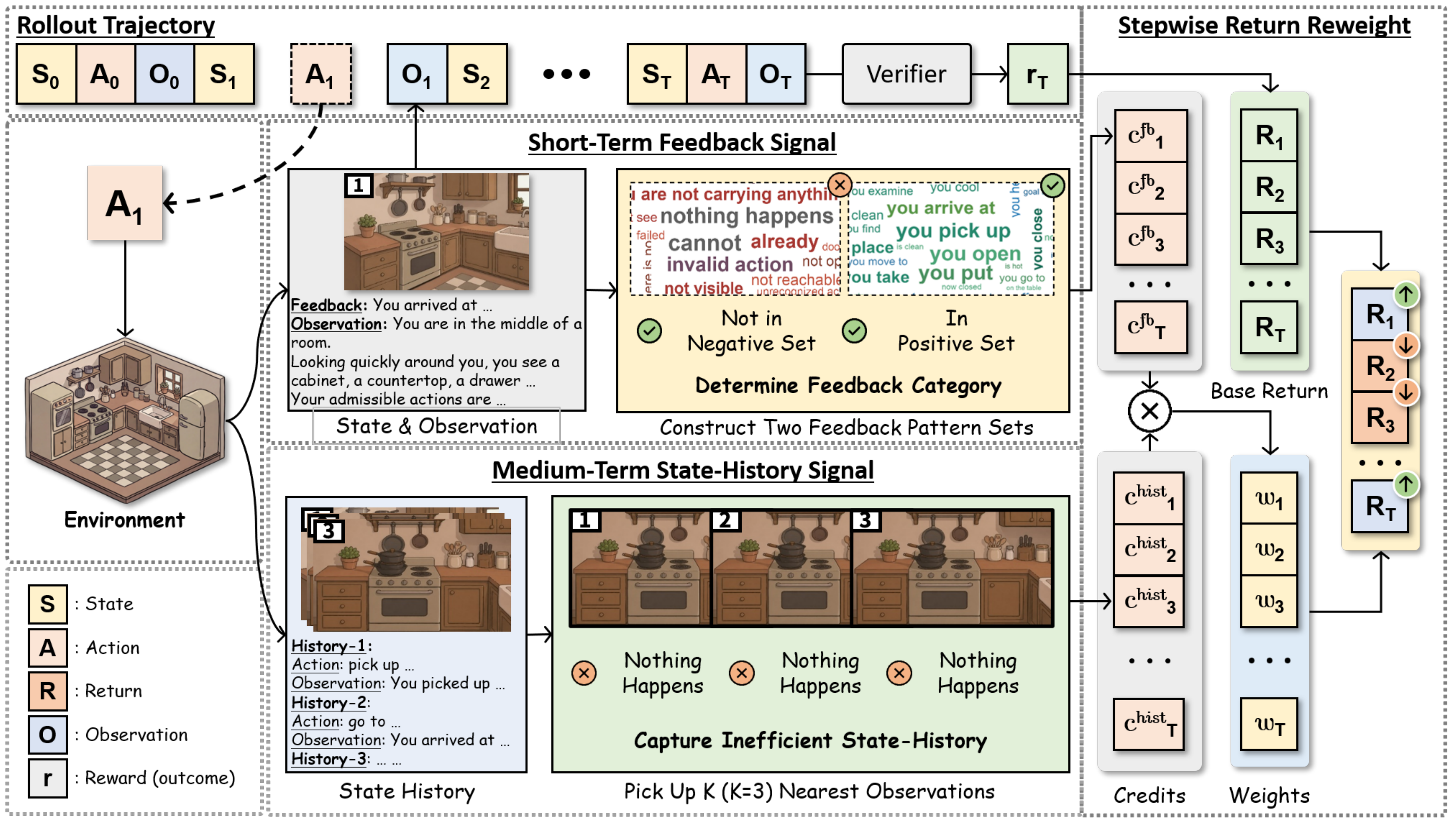}
  \caption{  Overview of EFCA. EFCA complements the long-term outcome signal with a short-term feedback credit
  from immediate environment responses and a medium-term state-history credit from recent action-
  feedback history. The combined credit score reweights the original step-level return, strengthening
  supervision for locally effective actions and suppressing repeated ineffective behaviors.}
  \label{fig:placeholder-pdf}
\end{figure*}

\subsection{Problem Formulation}
We consider long-horizon agentic tasks in which an LLM-based policy interacts with an environment over multiple steps. At step $t$, the agent observes a state-like context $o_t$, generates an action $a_t$, and receives environmental feedback $f_t$ together with an environment reward signal. 
A trajectory is therefore written as $\tau = \{(o_t, a_t, f_t, r_t)\}_{t=1}^{T}$. 
Following existing stepwise policy optimization methods, the policy is optimized at the level of intermediate steps rather than only at the trajectory level.

Let $R_t^{base}$ denote the original step-level return used by the underlying stepwise optimization algorithm, which is typically computed from discounted rewards. 
In most existing stepwise optimization methods, this return is constructed by backward discounted accumulation of trajectory-level rewards.
Formally, given a trajectory $\tau$ and a discount factor $\gamma \in [0,1]$, the base step-level return at step $t$ is written as

\begin{equation}
R_t^{\text{base}} = \sum_{t' = t}^{T} \gamma^{t' - t} r_{t'}.
\end{equation}

\noindent where $t'$ indexes the current and future steps along the same trajectory, ranging from the current step $t$ to the terminal step $T$. 
The term $r_{t'}$ denotes the reward received at step $t'$, while $\gamma^{t' - t}$ discounts rewards that occur farther in the future. 
Therefore, $R_t^{\text{base}}$ represents the discounted accumulation of all rewards from step $t$ onward, and serves as the standard step-level supervision signal in most existing stepwise optimization methods.
However, this discounting mechanism does not distinguish the contribution among different steps within a same trajectory, making the credit signals improperly assigned.

In this paper, our goal is to refine this supervision by introducing \emph{environment-grounded multi-timescale credit signals} that better reflect the local contribution and historical efficiency of each action. 
Importantly, we do not replace the underlying optimizer or introduce an explicit value critic; instead, we modify the step-level supervision signal in a plug-and-play manner.

% In this paper, our goal is to refine this supervision by introducing a \emph{environmental feedback-based customized credit signal} that better reflects the local contribution and historical efficiency of each action. Importantly, we do not replace the underlying optimizer or introduce an explicit value critic; instead, we modify the step-level supervision signal in a plug-and-play manner.

\subsection{Multi-timescale Credit Signal Construction}

The proposed credit signal is designed to complement the original discounted return, which mainly reflects the long-term outcome signal propagated from future rewards. 
Our key observation is that in long-horizon agentic tasks, whether an intermediate action deserves reinforcement depends not only on the eventual trajectory outcome, but also on finer-grained process evidence available during interaction. 
We therefore introduce two environment-grounded process signals. 
The \emph{short-term feedback signal} captures whether the current action has produced an immediate and meaningful local effect. 
The \emph{medium-term state-history signal} captures whether the recent interaction pattern indicates state-level progress or repeated ineffective behavior. 
Together, these signals provide a temporally structured view of credit assignment: the long-term outcome signal evaluates whether the task is completed, the short-term feedback signal evaluates whether the current action works, and the medium-term state-history signal evaluates whether recent behavior is progressing efficiently.

% Our customized credit signal is built from two complementary sources: immediate environmental feedback and repetition-aware historical failure patterns. 
% Intuitively, the first captures whether the current action appears locally useful, while the second captures whether the current action repeats behavior that has already been shown to be ineffective.

% (Add two examples of the failure modes associating to the two customized rewards)

% The proposed customized credit signal is designed to complement the original discounted return. 
% Our key observation is that in long-horizon agentic tasks, whether an intermediate action deserves reinforcement depends not only on the eventual trajectory outcome, but also on two finer-grained cues that are often available during interaction. 
% First, the immediate environmental feedback often reveals whether the current action has produced a meaningful local effect. 
% Second, the recent interaction history reveals whether the current action is merely repeating behavior that has already been shown to be ineffective. 
% These two sources capture different aspects of step quality: the former is \emph{locally outcome-aware}, while the latter is \emph{history-aware}.

Based on this intuition, we combine the two process signals into a unified step-level credit score. 
We denote the resulting customized credit score at step $t$ as

\begin{equation}
C_t = \alpha \cdot c_t^{\text{fb}} + \beta \cdot c_t^{\text{hist}},
\end{equation}

\noindent where $c_t^{\text{fb}}$ is the short-term feedback credit, $c_t^{\text{hist}}$ is the medium-term state-history credit, and $\alpha, \beta$ are corresponding coefficients controlling their relative importance. 
In this way, each component can be interpreted independently, ablated separately, and flexibly integrated into existing stepwise optimization pipelines. 
Intuitively, $C_t$ becomes larger when the current action appears locally useful and non-redundant, and becomes smaller when the action is associated with ineffective execution or repeated historical failure.

% \noindent where $c_t^{\text{fb}}$ is the feedback credit, $c_t^{rep}$ is the repetition credit, and $\alpha, \beta$ are corresponding coefficients controlling their relative importance. 
% This additive form is designed to keep the method simple and modular.
% In this way, each component can be interpreted independently, ablated separately, and flexibly integrated into existing stepwise optimization pipelines. 
% Intuitively, $C_t$ becomes larger when the current action appears locally useful and non-redundant, and becomes smaller when the action is associated with ineffective execution or repeated historical failure.

The combined score is then transformed into a step-dependent weight for return reweighting. 
In this way, the step-level supervision is adaptively calibrated according to both the immediate execution evidence and recent state-history evidence instead of being treated as uniformly reliable across all intermediate actions.

\subsection{Short-term Feedback Signal}
The short-term feedback signal extracts a local execution-quality signal from the immediate environment response. 
The key intuition is that the environment often directly exposes whether an action has produced a meaningful state transition. 
For example, in ALFWorld, responses such as \texttt{Nothing happens} usually indicate negligible contribution, while responses such as \texttt{You open ...}, \texttt{You pick up ...}, or \texttt{You put ...} often correspond to meaningful progress.

Formally, let $f_t$ be the feedback at step $t$, and let $\phi(\cdot)$ be a normalization function that lower-cases and standardizes the textual feedback. Let $\mathcal{P}$ and $\mathcal{N}$ denote positive and negative feedback pattern sets, respectively. The feedback credit is defined as
\begin{equation}
c_t^{\text{fb}} =
\begin{cases}
-1, & \text{if } \exists n \in \mathcal{N},\; n \preceq \phi(f_t), \\
+1, & \text{else if } \exists p \in \mathcal{P},\; p \preceq \phi(f_t), \\
0, & \text{otherwise.}
\end{cases}
\end{equation}
This design turns environment feedback into a simple and interpretable local step-level signal. The resulting credit serves as an auxiliary signal indicating whether the current action appears locally useful or locally ineffective.

\subsection{Medium-term State-History Signal}
The medium-term state-history signal captures short-horizon inefficiency from recent environmental
feedback.
Unlike the short-term feedback signal, which only evaluates the immediate effect of the current
action, the state-history signal examines whether the agent has been trapped in a locally
unproductive interaction pattern over the recent feedback window.

Let $c_i^{\text{fb}} \in \{-1,0,+1\}$ denote the short-term feedback credit at step $i$.
For step $t$, we define a recent feedback window of size $K$ as
\begin{equation}
\mathcal{W}_t^K = \{i \mid \max(1,t-K+1) \le i \le t\}.
\end{equation}
We then count the number of negative and positive feedback signals in this window:
\begin{equation}
N_t^- = \sum_{i \in \mathcal{W}_t^K} \mathbb{I}[c_i^{\text{fb}}=-1],
\quad
N_t^+ = \sum_{i \in \mathcal{W}_t^K} \mathbb{I}[c_i^{\text{fb}}=+1].
\end{equation}

The medium-term state-history credit is defined as
\begin{equation}
c_t^{\text{hist}} =
\begin{cases}
-1, & \text{if } |\mathcal{W}_t^K|=K \text{ and } N_t^- = K, \\
-\eta, & \text{else if } |\mathcal{W}_t^K|=K \text{ and } N_t^+ = 0, \\
0, & \text{otherwise,}
\end{cases}
\end{equation}
where $\eta \in (0,1)$ is a mild stagnation penalty.

This formulation identifies two common low-efficiency patterns in long-horizon agentic tasks.
The first case captures consecutive negative feedback, indicating that recent actions repeatedly
fail or violate environmental constraints.
The second case captures consecutive absence of positive feedback, indicating that the agent has
made no explicit local progress over the recent window.
In both cases, the history signal provides a medium-term correction to the original step-level
return without introducing positive rewards or an additional learned evaluator.

\subsection{Return Reweighting with Multi-timescale Signals}
The two process credit components are integrated with the long-term outcome signal through a return reweighting mechanism. 
Instead of directly replacing the base return, we compute a step-dependent weight and apply it to the original step-level return. 
Specifically, we define
\begin{equation}
w_t = \mathrm{clip}(1 + \lambda C_t, w_{min}, w_{max}),
\end{equation}
where $\lambda$ controls the overall reweighting strength, and $w_{min}, w_{max}$ are clipping bounds for stability.
The final reweighted return is then
\begin{equation}
\tilde{R}_t = w_t \cdot R_t^{base}.
\end{equation}
This mechanism has two advantages. 
First, it preserves compatibility with existing stepwise optimization methods such as GiGPO and HGPO, because the underlying optimization pipeline remains unchanged. 
Second, it provides a direct way to inject richer step-level supervision without introducing an explicit critic.

\section{Experiments}
\begin{table*}[t]
    \centering
    \resizebox{0.99\linewidth}{!}{\input{tables/main_table}}
    \caption{Main results on ALFWorld and WebShop. Results marked with $\dagger$ are taken from the original GraphGPO paper, and WebShop results marked with $\ddagger$ are taken from the original HGPO paper.}
    \label{tab:main_results}
\end{table*}
\subsection{Experimental Setup}

\subsubsection{Evaluation Environments.}
We conduct training on two challenging benchmarks, ALFWorld and WebShop  \cite{Shridhar2020ALFWorldAT,Yao2022WebShopTS}.
ALFWorld serves as an embodied environment for assessing the ability of LLM agents to carry out multi-step decision-making. 
In each episode, the agent is given a textual goal and must fulfill it through multi-turn interactions with the environment. 
The benchmark consists of 3,827 task instances covering six categories of common household activities: Pick \& Place (Pick), Examine in Light (Look), Clean \& Place (Clean), Heat \& Place (Heat), Cool \& Place (Cool), and Pick Two \& Place (Pick2).
WebShop is a web-based interactive environment designed to evaluate LLM agents under realistic online shopping scenarios. 
To accomplish a given task, the agent must interact with a simulated HTML-based shopping website to identify, navigate to, and purchase an appropriate item. 
With over 1.1 million products and 12k user instructions, WebShop provides a rich and diverse action space.
During evaluation, each environment is run with three different random seeds, and we report the mean and standard deviation of performance. 
The maximum number of interaction steps is set to 50 for ALFWorld and 30 for WebShop. 

\subsubsection{Baseline Methods.}
For LLM-based agents, we compare EFCA with a diverse set of competitive baselines. 
These baselines can be grouped into three categories: closed-source LLMs, prompting-based agent paradigms, and RL-based training methods.
Specifically, for closed-source models, we include GPT-4o and Gemini 2.5 Pro as strong general-purpose baselines \cite{Hurst2024GPT4oSC,Reid2024Gemini1U}. 
For prompting-based methods, we consider ReAct and Reflexion, both of which rely exclusively on in-context information to solve multi-turn tasks without parameter updates; we additionally include a reasoning-free baseline for comparison \cite{yao2022react,Shinn2023ReflexionLA}. 
For RL-based methods, we adopt GRPO as a representative group-based trajectory-level optimization approach \cite{Schulman2017ProximalPO,Shao2024DeepSeekMathPT}.
Furthermore, we compare EFCA with several stepwise policy optimization methods, including GiGPO, GraphGPO, and HGPO \cite{Feng2025GroupinGroupPO,Cheng2026BeyondTA,He2026HierarchyofGroupsPO}. 
To simplify implementation and maintain consistency with prior work, we directly reuse part of the reported results from the corresponding original papers.

\subsubsection{Implementation Details.}
Following prior work, we adopt Qwen2.5-1.5B-Instruct and Qwen2.5-7B-Instruct as our backbone models \cite{Yang2024Qwen25TR}. 
To maintain consistency with existing agent frameworks, the agent retains only the two most recent interaction steps as memory while discarding earlier history. 
Moreover, the decision-making protocol follows the ReAct setting, where at each step the LLM agent is prompted to first generate a reasoning trace enclosed within \texttt{<think></think>} tags and then produce a textual action enclosed within \texttt{<action></action>} tags.
To ensure fair comparison, all RL-based methods share the same hyperparameter configuration; in particular, the rollout group size $N$ is set to 8. 
For simplicity, we construct the feedback set by directly extracting the feedback signals provided by the environment.

\subsection{Experimental Results}

\begin{figure*}[t]
\centering
\input{figures/ablation}
\caption{Ablation results of EFCA on ALFWorld under different hyperparameter settings. The
three subfigures show the effect of the overall return reweighting strength, the medium-
term state-history coefficient, and the short-term feedback coefficient, respectively.}
\label{fig:ablation_multi_timescale}
\end{figure*}
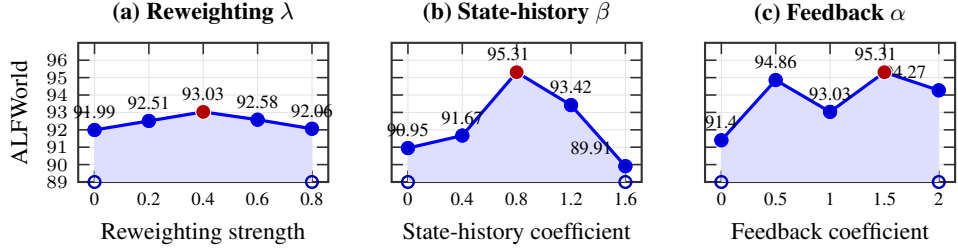

\subsubsection{Performance on Agentic Benchmarks.}
Table~\ref{tab:main_results} reports the main results on ALFWorld and WebShop. 
Overall, EFCA consistently improves upon the original HGPO baseline and remains competitive with strong critic-free stepwise RL methods, including GiGPO and GraphGPO. 
The gains are most pronounced on ALFWorld, where EFCA achieves the best overall score under both model scales. 
With Qwen2.5-1.5B-Instruct, EFCA reaches an overall ALFWorld score of \textbf{95.31}, outperforming HGPO by $+3.32$ points, GiGPO by $+4.43$ points, and GraphGPO by $+2.60$ points. 
With Qwen2.5-7B-Instruct, EFCA further improves the overall ALFWorld score to \textbf{96.03}, exceeding HGPO by $+4.04$ points, GiGPO by $+1.76$ points, and GraphGPO by $+0.72$ points. 
These results suggest that EFCA provides a stable and effective refinement of step-level credit signals across both small and larger backbones.

% A closer look at the per-task breakdown shows that EFCA is particularly effective on ALFWorld sub-tasks where repeated low-contribution actions are more likely to hinder long-horizon planning. Under the 1.5B backbone, EFCA substantially improves over HGPO on \textit{Cool} (94.39 vs.~85.63) and \textit{Pick2} (93.18 vs.~82.57), while remaining highly competitive on \textit{Pick} and \textit{Clean}. Under the 7B backbone, EFCA achieves strong and balanced performance across nearly all ALFWorld categories, including 99.56 on \textit{Clean}, 96.40 on \textit{Heat}, 92.45 on \textit{Cool}, and 94.14 on \textit{Pick2}. This pattern is consistent with the design of EFCA: feedback credit assignment captures immediate execution quality, repetition credit assignment suppresses repeated failed behaviors, and return reweighting injects both signals into the original discounted return without changing the underlying critic-free optimization pipeline.

EFCA also shows a distinct advantage on WebShop when evaluated by \emph{Task Score}. 
Specifically, EFCA achieves the best Task Score under both backbones, reaching \textbf{89.81} with Qwen2.5-1.5B-Instruct and \textbf{89.06} with Qwen2.5-7B-Instruct, outperforming all compared baselines, including GraphGPO. 
This result is particularly meaningful because WebShop Task Score is designed to measure how well the purchased item matches the instruction-specified product requirements, considering attributes, options, type, and price, rather than merely whether the agent executes a purchase action at the end of the trajectory. 
In other words, EFCA improves not only action completion, but also task-level alignment between the agent's final purchase and the user's intended product specification. 
By contrast, although GraphGPO attains slightly higher WebShop task success rate, EFCA's stronger Task Score suggests better semantic adaptation to the shopping objective itself, indicating that its credit assignment mechanism more effectively guides the policy toward purchasing \emph{better-matched} products rather than simply completing a purchase outcome.

\subsection{Training Dynamics of EFCA.}

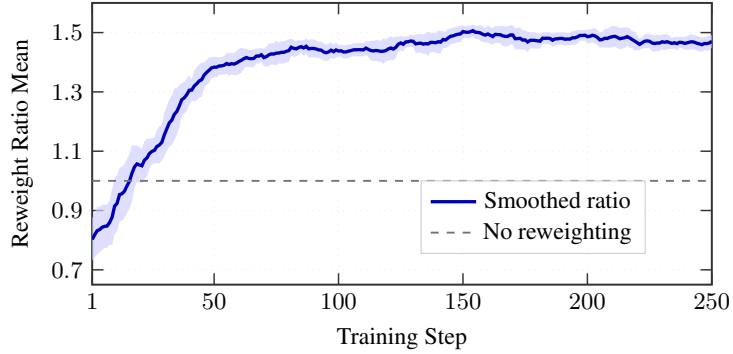
\begin{wrapfigure}{r}{0.7\textwidth}
   \input{figures/debug_step_reweight_ratio_mean_pgfplots}
   \caption{Mean reweighting ratio over training steps. The shaded area
   shows the raw-value band, while the solid line shows the smoothed
   trend. The dashed line indicates the no-reweighting baseline where
   the ratio equals 1.0.}
   \label{fig:debug_step_reweight_ratio_mean}
 \end{wrapfigure}

Figure~\ref{fig:debug_step_reweight_ratio_mean}
illustrates the evolution of the mean reweighting ratio
during training.
At the early stage, the ratio is mostly below or close to
1.0, indicating that EFCA initially assigns conservative
weights to intermediate actions when the policy is still
unstable and many actions receive weak or negative process
evidence.
As training proceeds, the mean ratio gradually increases
and remains consistently above the no-reweighting
baseline, eventually stabilizing around 1.4--1.5.
This trend suggests that the policy increasingly produces
actions that are supported by positive short-term feedback
and medium-term state-history signals.
In other words, EFCA does not simply impose a fixed global
scaling on all steps; instead, it adaptively calibrates
the step-level return according to the quality of
environment-grounded process evidence.
The stable ratio above 1.0 in later training indicates
that more intermediate actions are recognized as locally
effective or historically efficient, leading to stronger
supervision for useful behaviors while still preserving
the original long-term outcome signal.

\subsubsection{Comparison of Stepwise Return Variance.}

\begin{wrapfigure}{R}{0.7\textwidth}
    \input{figures/debug_base_step_return_std_pgfplots}
    \caption{Standard deviation of base and reweighted step returns over
    training steps. The shaded bands show raw-value variations around the
    smoothed trends, while the solid lines show the smoothed curves.}
    \label{fig:debug_base_step_return_std}
\end{wrapfigure}
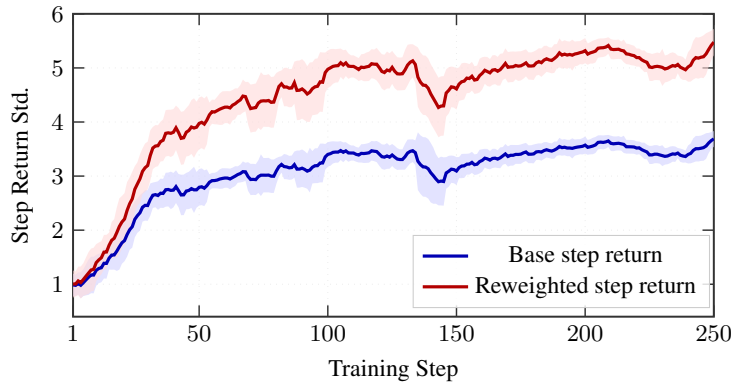
  
Figure~\ref{fig:debug_base_step_return_std} compares the
standard deviation of the original step returns and the
reweighted step returns during training.
The reweighted returns consistently exhibit a larger
standard deviation than the base returns, especially in
the later training stage.
This indicates that EFCA does not uniformly scale all
intermediate steps, but instead assigns differentiated
weights according to environment-grounded process
evidence.
By incorporating short-term feedback and medium-term
state-history signals, EFCA more effectively distinguishes
high-contribution actions from invalid, redundant, or low-
contribution actions.
As a result, steps within the same trajectory receive more
diverse supervision strengths, leading to a larger
variance in the reweighted step-level targets.
This larger variance therefore reflects more
discriminative credit assignment rather than instability,
showing that EFCA can better capture heterogeneous step
contributions in long-horizon agentic tasks.

\subsubsection{Ablation Study.}

We conduct controlled ablations to examine the contribution of the two process signals in EFCA. 
As shown in Figure~\ref{fig:ablation_multi_timescale}, the full model reaches an ALFWorld score of \textbf{95.31}. 
Removing the medium-term state-history signal lowers the score to 90.95 ($-4.36$), showing that recent interaction history helps detect repeated ineffective actions. Disabling the short-term feedback signal reduces the score to 91.40 ($-3.91$), indicating that immediate environmental feedback is also important for judging local progress. 
Overall, combining both signals works better than using either one alone.

We further study three key hyperparameters. 
For the overall return reweighting strength, performance rises from 91.99 without reweighting to 92.51, 93.03, 92.58, and 92.06 as the value increases from 0.2 to 0.8, with the best result at 0.4. 
This suggests that moderate reweighting is helpful, while overly strong reweighting may introduce noise.

For the medium-term state-history coefficient, increasing it from 0 to 0.8 improves performance from 90.95 to \textbf{95.31}, confirming the value of suppressing repeated low-contribution actions. 
However, larger values of 1.2 and 1.6 reduce the score to 93.42 and 89.91, which suggests that this signal should remain a calibrated correction rather than a dominant penalty.

For the short-term feedback coefficient, all nonzero settings outperform the no-feedback score of 91.40.
The scores are 94.86, 93.03, 95.31, and 94.27 for coefficients 0.5, 1.0, 1.5, and 2.0. 
The best result is achieved at 1.5, showing that short-term feedback is robust and useful for step-level credit assignment.

\section{Related Work}

In recent years, LLMs have served as the primary backbone for text-driven agent tasks \cite{Schick2023ToolformerLM,Gur2023ARW,Chen2025ReinforcementLF}. 
Adapting LLM-based agents to specific environments has proven highly effective in enhancing their performance \cite{Wang2023AdaptingLA,Chen2025TestTimeAF,Yu2026ReinforcementWM}. 
However, prior works mainly focus on optimizing these models using sparse outcome rewards, largely neglecting the abundant process signals implicitly contained within environmental feedback \cite{Lattimer2024SparseRC,Yuan2026VerifiablePR}.

\subsection{Large Language Model-based Agent}
Large language models are increasingly being deployed as general-purpose agents that can interact with external environments, utilize tools, and solve tasks through multi-step decision-making   \cite{Schick2023ToolformerLM,Wang2023VoyagerAO,Zhou2023SOTOPIAIE}. 
Unlike traditional NLP paradigms, which merely require the generation of a single static output, LLM-based agents must operate within a closed-loop interactive framework. 
They are required to repeatedly observe environmental states, execute actions, process feedback, and sequentially adapt their strategies. 
This shift has catalyzed extensive research across tool-use agents, embodied agents, and web agents, highlighting a fundamental transition: the core challenge is no longer merely reasoning over static text, but rather reasoning dynamically through continuous interaction   \cite{Li2024EmbodiedAI,Furuta2023MultimodalWN,Gur2023ARW}.

These agentic tasks generally exhibit several intrinsic properties   \cite{Shridhar2020ALFWorldAT,Yao2022WebShopTS,Chen2025ReinforcementLF}. 
First, they operate under partial observability, where environment dynamics must be inferred through opaque textual or multimodal feedback. 
Second, achieving the task objective necessitates sequential, multi-step coordination rather than a solitary output generation. 
Third, the marginal contribution of intermediate actions is highly skewed: while specific actions are critical bottlenecks for progress, others remain transitional or entirely superfluous. 
Together, these properties make the optimization of LLM-based agents an exceptionally promising yet fundamentally challenging domain for reinforcement learning \cite{Du2025ASO,Chen2025ReinforcementLF}.

\subsection{Adapting LLMs to Agentic Tasks}
% Although the representation issue is mostly solved, the adaption still remains as a problem.
% To mitigate this problem, training-free methods are firstly considered, such as ReAct and Pre-Act, in that the LLMs are prompted to generate think traces before taking any action.

% (ReAct)
% (RL)
% (Verl-Agent)
Existing approaches for adapting LLMs to agentic tasks can be broadly categorized into prompting-based and optimization-based methods \cite{Du2025ASO}. 
Prompting-based techniques seek to enhance agent behavior by engineering sophisticated interaction templates, reasoning traces, planning structures, or memory modules   \cite{yao2022react,Shinn2023ReflexionLA,Erdogan2025PlanandActIP}. 
While representative works demonstrate that prompting can significantly boost short-horizon decision quality and tool-use proficiency without parameter update, these methods face inherent limitations. 
As they rely entirely on in-context behavior shaping rather than direct parameter optimization, they frequently struggle to generalize to long-horizon tasks that demand complex error recovery and operate under sparse supervision \cite{Kim2026OnTL,Chen2025IterResearchRL}.

In contrast, optimization-based methods adapt LLMs to agentic tasks via direct parameter updates, utilizing techniques such as supervised fine-tuning on expert trajectories or reinforcement learning from interactive rollouts \cite{Wang2023AdaptingLA,Chen2025ReinforcementLF,Yu2026ReinforcementWM}. 
Compared to prompting, these approaches enable the model to learn more stable and robust behavioral patterns directly from experience \cite{Wang2025MROER}. 
Within this category, a critical paradigm shift is the transition from trajectory-wise to stepwise optimization, where the model is trained at the granularity of intermediate actions rather than complete episodes \cite{Wang2026StepPOSP,Wang2025SPARLRL,Feng2025GroupinGroupPO}. 
While this shift is particularly crucial for mastering long-horizon tasks, it concurrently elevates the necessity of designing highly accurate step-level credit assignment signals   \cite{Pignatelli2023ASO,Tan2026HindsightCA,Cheng2026BeyondTA}.

\section{Conclusion}
In this work, we present Environmental Feedback-based Credit Assignment (EFCA), a lightweight multi-timescale credit assignment method for training LLM agents with real environment feedback.
In contrast to existing methods that perform trajectory-level attribution based primarily on final outcomes, EFCA transforms environmental feedback into process-level supervision and assigns credit at different timescale according to local progress toward the overall task objective.
Through extensive experiments on a diverse set of multi-turn agentic benchmarks, we show that EFCA consistently surpasses prior group-based methods, benefiting from its ability to exploit environmental feedback for more effective step-level credit assignment.
These results suggest that EFCA serves as a practical complement to existing group-based RL methods and offers a promising direction for improving credit assignment in long-horizon agentic tasks.

\bibliography{main}
\bibliographystyle{iclr2027_conference}

\newpage

\appendix

\section{Additional Method Details}

\subsection{Overview of EFCA}
Environmental Feedback-based Credit Assignment (EFCA) refines the step-level return used by critic-free stepwise policy optimization.  Given a trajectory
\begin{equation}
\tau = \{(o_t,a_t,f_t,r_t)\}_{t=1}^{T},
\end{equation}
where $o_t$ is the observation or state-like context, $a_t$ is the action, $f_t$ is the environment feedback, and $r_t$ is the scalar reward, the base return is computed as
\begin{equation}
R_t^{\mathrm{base}} = \sum_{t'=t}^{T} \gamma^{t'-t} r_{t'}.
\end{equation}
EFCA does not replace this return.  Instead, it constructs environment-grounded process credit signals and uses them to reweight $R_t^{\mathrm{base}}$:
\begin{equation}
\widetilde{R}_t = w_t R_t^{\mathrm{base}}.
\end{equation}
The central design goal is to preserve compatibility with existing stepwise optimization pipelines while making the step-level supervision more discriminative.

\subsection{Short-term Feedback Credit}
The short-term feedback credit measures the immediate execution effect of the current action.  Let $\phi(\cdot)$ denote a normalization function that lower-cases feedback text, removes excessive whitespace, and standardizes surface variants.  Let $\mathcal{P}$ and $\mathcal{N}$ denote the positive and negative feedback pattern sets.  The feedback credit is defined as
\begin{equation}
 c_t^{\mathrm{fb}} =
 \begin{cases}
 -1, & \text{if } \exists n \in \mathcal{N},\; n \preceq \phi(f_t), \\
 +1, & \text{else if } \exists p \in \mathcal{P},\; p \preceq \phi(f_t), \\
 0, & \text{otherwise,}
 \end{cases}
\end{equation}
where $x \preceq y$ means that pattern $x$ is matched in normalized text $y$.

The negative set $\mathcal{N}$ contains feedback patterns indicating invalid, ineffective, or redundant execution.  In ALFWorld-style environments, examples include \texttt{nothing happens}, \texttt{you are not carrying anything}, \texttt{cannot}, \texttt{not valid}, and \texttt{already}.  The positive set $\mathcal{P}$ contains feedback patterns indicating explicit local progress, such as \texttt{you open}, \texttt{you pick up}, \texttt{you put}, \texttt{you clean}, \texttt{you heat}, and \texttt{you cool}.  For WebShop-style environments, the pattern sets are adapted to navigation and purchase feedback, including successful search, item selection, option selection, and purchase-related state changes.

\subsection{Medium-term State-history Credit}
The medium-term state-history credit captures whether recent interactions indicate local stagnation or repeated ineffective behavior.  Unlike the short-term credit, which only evaluates the current step, this signal aggregates feedback evidence over a recent window.

Let $K$ be the history window size.  For step $t$, define
\begin{equation}
\mathcal{W}_t^K = \{i \mid \max(1,t-K+1) \leq i \leq t\}.
\end{equation}
The numbers of negative and positive feedback signals in this window are
\begin{equation}
N_t^- = \sum_{i \in \mathcal{W}_t^K} \mathbb{I}[c_i^{\mathrm{fb}}=-1],
\quad
N_t^+ = \sum_{i \in \mathcal{W}_t^K} \mathbb{I}[c_i^{\mathrm{fb}}=+1].
\end{equation}
The state-history credit is then
\begin{equation}
 c_t^{\mathrm{hist}} =
 \begin{cases}
 -1, & \text{if } |\mathcal{W}_t^K|=K \text{ and } N_t^- = K, \\
 -\eta, & \text{else if } |\mathcal{W}_t^K|=K \text{ and } N_t^+ = 0, \\
 0, & \text{otherwise,}
 \end{cases}
\end{equation}
where $\eta \in (0,1)$ is a mild stagnation penalty.  The first case detects consecutive negative feedback.  The second case detects a recent window with no explicit positive progress.  Both cases are designed as negative corrections rather than additional positive rewards, which helps avoid over-shaping the environment objective.

\subsection{Combined Credit and Return Reweighting}
EFCA combines the short-term and medium-term process signals into a unified credit score:
\begin{equation}
C_t = \alpha c_t^{\mathrm{fb}} + \beta c_t^{\mathrm{hist}},
\end{equation}
where $\alpha$ and $\beta$ control the relative strengths of the feedback and state-history signals.  The score is transformed into a reweighting coefficient:
\begin{equation}
w_t = \mathrm{clip}(1 + \lambda C_t, w_{\min}, w_{\max}),
\end{equation}
where $\lambda$ controls the overall reweighting strength and $w_{\min}, w_{\max}$ prevent extreme updates.  The final training target is
\begin{equation}
\widetilde{R}_t = w_t R_t^{\mathrm{base}}.
\end{equation}
When $C_t>0$, EFCA strengthens the return for steps supported by positive process evidence.  When $C_t<0$, EFCA suppresses the return for steps associated with invalid execution, ineffective behavior, or short-horizon stagnation.

\subsection{Pseudocode}
Algorithm~\ref{alg:efca} summarizes the EFCA credit construction and return reweighting procedure.

\begin{algorithm*}[t]
\caption{Environmental Feedback-based Credit Assignment}
\label{alg:efca}
\begin{algorithmic}[1]
\REQUIRE Trajectory $\tau=\{(o_t,a_t,f_t,r_t)\}_{t=1}^{T}$; discount factor $\gamma$; feedback pattern sets $\mathcal{P},\mathcal{N}$; window size $K$; coefficients $\alpha,\beta,\lambda$; clipping bounds $w_{\min},w_{\max}$; stagnation penalty $\eta$.
\ENSURE Reweighted step returns $\{\widetilde{R}_t\}_{t=1}^{T}$.
\FOR{$t=1$ to $T$}
    \STATE Compute $R_t^{\mathrm{base}} = \sum_{t'=t}^{T} \gamma^{t'-t} r_{t'}$.
    \STATE Normalize feedback $\bar{f}_t \leftarrow \phi(f_t)$.
    \IF{$\exists n\in\mathcal{N}$ such that $n \preceq \bar{f}_t$}
        \STATE $c_t^{\mathrm{fb}} \leftarrow -1$.
    \ELSIF{$\exists p\in\mathcal{P}$ such that $p \preceq \bar{f}_t$}
        \STATE $c_t^{\mathrm{fb}} \leftarrow +1$.
    \ELSE
        \STATE $c_t^{\mathrm{fb}} \leftarrow 0$.
    \ENDIF
\ENDFOR
\FOR{$t=1$ to $T$}
    \STATE Construct $\mathcal{W}_t^K = \{i \mid \max(1,t-K+1) \leq i \leq t\}$.
    \STATE Compute $N_t^- = \sum_{i \in \mathcal{W}_t^K}\mathbb{I}[c_i^{\mathrm{fb}}=-1]$ and $N_t^+ = \sum_{i \in \mathcal{W}_t^K}\mathbb{I}[c_i^{\mathrm{fb}}=+1]$.
    \IF{$|\mathcal{W}_t^K|=K$ and $N_t^- = K$}
        \STATE $c_t^{\mathrm{hist}} \leftarrow -1$.
    \ELSIF{$|\mathcal{W}_t^K|=K$ and $N_t^+=0$}
        \STATE $c_t^{\mathrm{hist}} \leftarrow -\eta$.
    \ELSE
        \STATE $c_t^{\mathrm{hist}} \leftarrow 0$.
    \ENDIF
    \STATE $C_t \leftarrow \alpha c_t^{\mathrm{fb}} + \beta c_t^{\mathrm{hist}}$.
    \STATE $w_t \leftarrow \mathrm{clip}(1+\lambda C_t,w_{\min},w_{\max})$.
    \STATE $\widetilde{R}_t \leftarrow w_t R_t^{\mathrm{base}}$.
\ENDFOR
\RETURN $\{\widetilde{R}_t\}_{t=1}^{T}$.
\end{algorithmic}
\end{algorithm*}

\section{Implementation Details}

\subsection{Agent Interaction Protocol}
We follow a ReAct-style interaction protocol.  At each environment step, the model first generates a reasoning trace enclosed by \texttt{<think>} and \texttt{</think>} tags, and then generates an executable action enclosed by \texttt{<action>} and \texttt{</action>} tags.  The environment executes the action and returns textual feedback.  EFCA only uses this feedback and the original reward signal; it does not require preference annotations, learned reward models, or a value critic.

\subsection{Memory and Rollout Settings}
For consistency with the main experiments, the agent keeps only the two most recent interaction steps in its prompt memory.  Earlier observations and feedback are discarded from the model input, but EFCA can still compute credit using the recorded trajectory logs during training.  The maximum number of interaction steps is set to 50 for ALFWorld and 30 for WebShop.  For RL-based baselines and EFCA, the rollout group size is set to $N=8$.

\subsection{Hyperparameters}
Table~\ref{tab:appendix_hyperparams} summarizes the main EFCA hyperparameters.  The tested ranges correspond to the ablation dimensions discussed in the main paper.  Exact values can be adjusted for a specific optimizer, but the important principle is to keep the reweighting moderate and bounded.

\begin{table*}[t]
\centering
\small
\begin{tabular}{lll}
\toprule
Hyperparameter & Meaning & Typical / tested values \\
\midrule
$K$ & State-history window size & $2,3,5$ \\
$\eta$ & Mild stagnation penalty & $0.5$ \\
$\alpha$ & Feedback-credit coefficient & $0.5,1.0,1.5,2.0$ \\
$\beta$ & State-history coefficient & $0,0.4,0.8,1.2,1.6$ \\
$\lambda$ & Reweighting strength & $0,0.2,0.4,0.6,0.8$ \\
$w_{\min}$ & Minimum weight & $0.2$ or $0.3$ \\
$w_{\max}$ & Maximum weight & $1.8$ or $2.0$ \\
\bottomrule
\end{tabular}
\caption{Main EFCA hyperparameters and tested ranges.}
\label{tab:appendix_hyperparams}
\end{table*}

\subsection{Pattern Construction}
The feedback pattern sets are constructed from environment-provided messages rather than from model-generated text.  This design keeps the credit signal grounded in the environment.  In practice, pattern construction follows three steps: normalize environment feedback, collect frequent messages that clearly indicate progress or failure, and assign them to $\mathcal{P}$ or $\mathcal{N}$.  Ambiguous messages are left unmatched and therefore receive zero feedback credit.  This conservative strategy reduces the risk of injecting noisy credit.

\begin{figure*}[t]
\centering
\begin{minipage}{0.48\textwidth}
    \centering
    \includegraphics[width=\linewidth]{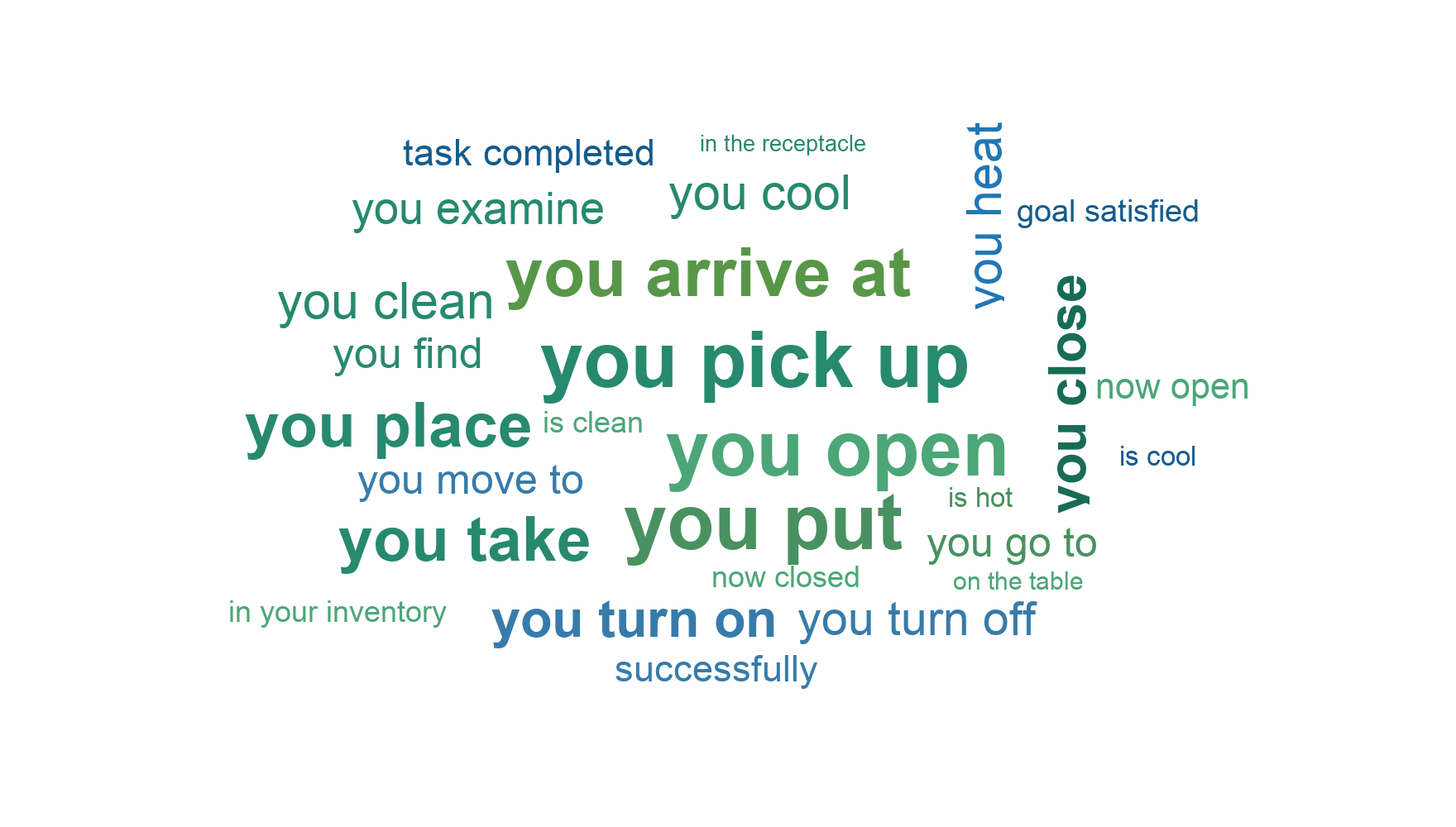}
    \par\smallskip
    {\small (a) Positive feedback patterns}
\end{minipage}
\hfill
\begin{minipage}{0.48\textwidth}
    \centering
    \includegraphics[width=\linewidth]{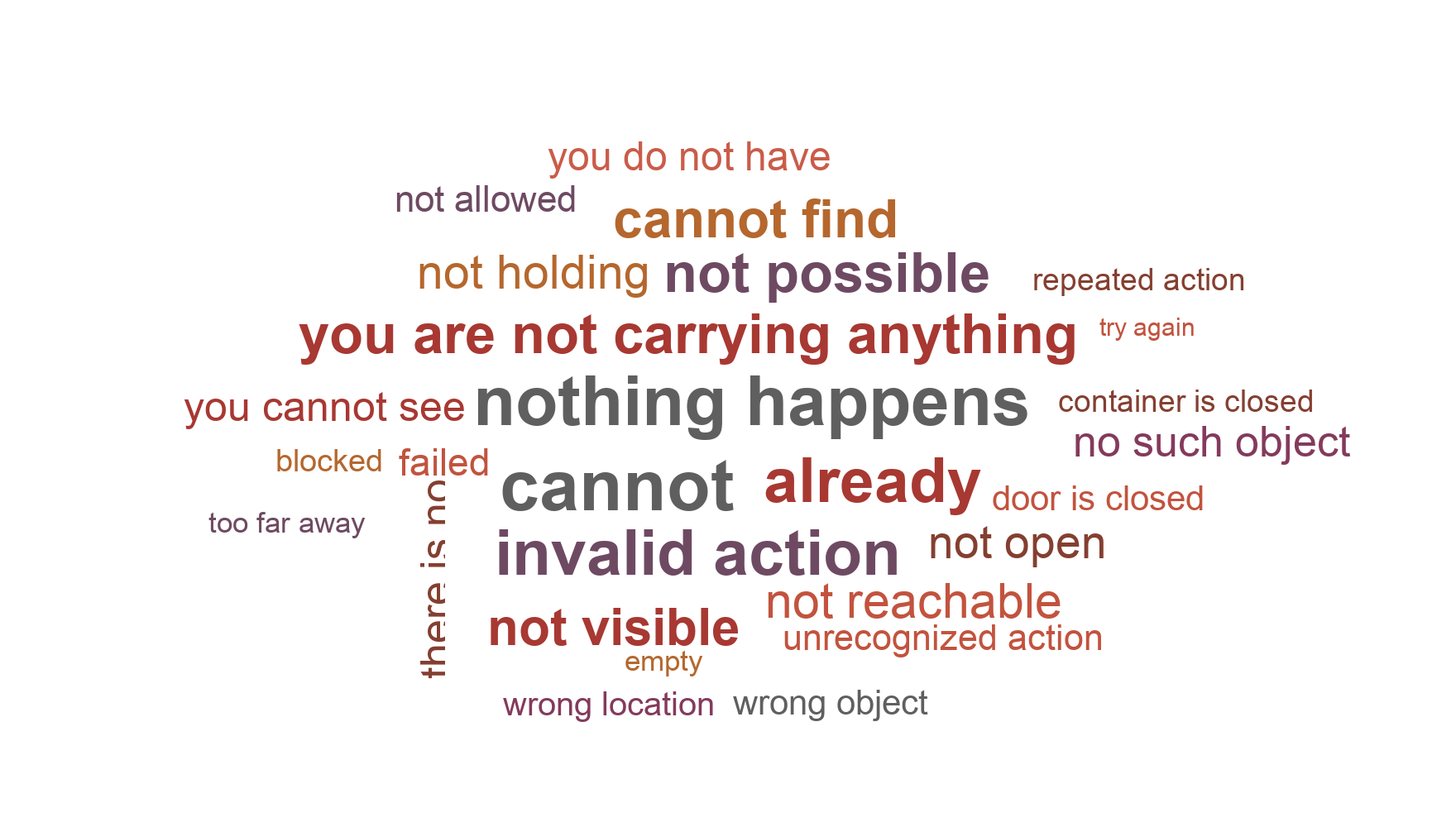}
    \par\smallskip
    {\small (b) Negative feedback patterns}
\end{minipage}
\caption{Word clouds of ALFWorld feedback patterns used to construct the positive set $\mathcal{P}$ and negative set $\mathcal{N}$. Larger phrases correspond to more representative feedback patterns observed in the interaction logs. The positive set captures explicit local progress, while the negative set captures invalid, infeasible, or ineffective interactions.}
\label{fig:alfworld_feedback_wordcloud}
\end{figure*}

Figure~\ref{fig:alfworld_feedback_wordcloud} illustrates this construction process using ALFWorld as an example.  The positive pattern set $\mathcal{P}$ mainly contains phrases that indicate explicit state progress, such as reaching a location, opening or closing an object, picking up an item, placing an item, or completing a task-relevant transformation.  The negative pattern set $\mathcal{N}$ mainly contains phrases that indicate invalid execution, impossible actions, missing objects, unreachable targets, empty inventory, repeated actions, or unchanged states.  These two sets therefore provide a simple environment-grounded interface for converting textual feedback into short-term credit signals.

\subsection{Computational Cost}
EFCA is designed as a lightweight return reweighting module.  It does not introduce an additional reward model, value critic, or environment-specific neural scorer.  The extra computation mainly consists of normalizing textual feedback, matching it against the positive and negative pattern sets, maintaining a short feedback-history window, and applying a scalar reweighting coefficient to the base step return.  These operations are negligible compared with rollout generation and policy optimization for the backbone LLM.  Therefore, the practical computational cost is dominated by the backbone scale and the underlying RL training pipeline rather than by EFCA itself.

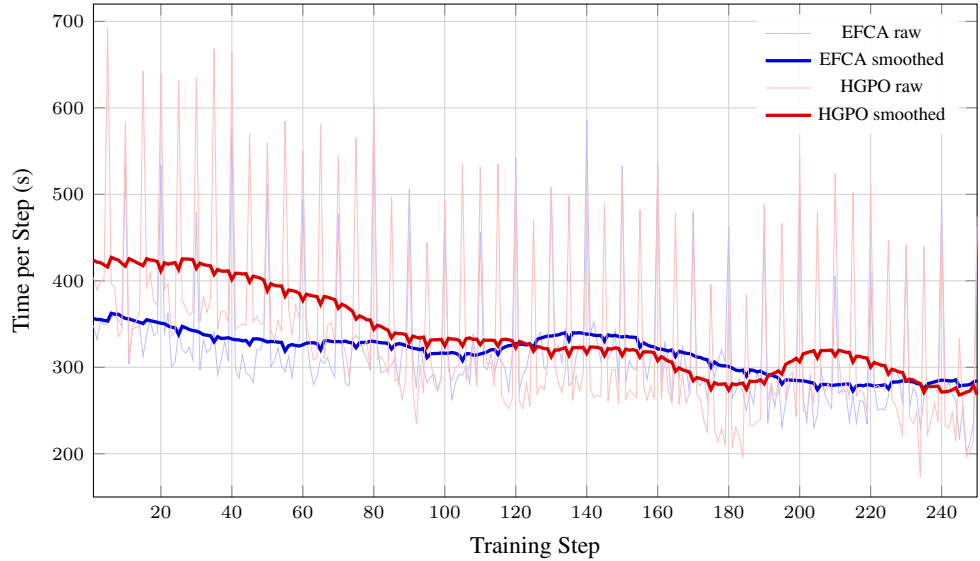
\begin{figure*}[t]
\centering
\input{figures/computational_cost_pgfplots}
\caption{Wall-clock time per training step for EFCA and HGPO.  Thin lines show raw step-level timing, while thick lines show the smoothed trend.  EFCA has lower average per-step cost over the full run, suggesting that feedback matching and return reweighting add negligible overhead compared with rollout generation and policy optimization.}
\label{fig:computational_cost_timing}
\end{figure*}

Figure~\ref{fig:computational_cost_timing} compares the wall-clock time per training step of EFCA and HGPO over 250 training steps.  EFCA takes 317.05 seconds per step on average, while HGPO takes 338.58 seconds per step, corresponding to an average reduction of 21.53 seconds per step, or 6.36\%.  Accumulated over the 250-step run, EFCA requires 22.02 wall-clock hours compared with 23.51 hours for HGPO, saving about 1.50 hours.  The reduction is most visible in the early stage, where EFCA is 17.21\% faster over the first 50 steps.  In the middle stage the two methods have similar cost, and in the last 50 steps EFCA remains 5.30\% faster on average.  These results indicate that the proposed feedback-based reweighting does not introduce a measurable runtime burden; instead, the observed cost remains comparable to or lower than the HGPO baseline under the same training pipeline.

Table~\ref{tab:compute_cost} reports the GPU configuration used in our experiments.  Training the Qwen2.5-1.5B-Instruct backbone requires 2 H100 GPUs, while evaluation uses 1 H100 GPU.  Training the Qwen2.5-7B-Instruct backbone requires 4 H100 GPUs, while evaluation uses 2 H100 GPUs.  This scaling is consistent with the larger memory and throughput requirements of the 7B backbone.  Since EFCA only modifies step-level return construction, it can be added to the same training and evaluation setup without requiring separate model inference beyond the policy rollouts.

\begin{table}[t]
\centering
\small
\begin{tabular}{lcc}
\toprule
Backbone & Training & Evaluation \\
\midrule
Qwen2.5-1.5B-Instruct & 2 $\times$ H100 & 1 $\times$ H100 \\
Qwen2.5-7B-Instruct & 4 $\times$ H100 & 2 $\times$ H100 \\
\bottomrule
\end{tabular}
\caption{GPU configurations used for training and evaluation. EFCA itself only adds lightweight feedback matching and return reweighting, so the computational cost is dominated by the backbone LLM and rollout generation.}
\label{tab:compute_cost}
\end{table}

\section{Additional Experimental Details}

\subsection{ALFWorld}
ALFWorld evaluates embodied text-based decision making.  Each task requires the agent to complete a household goal through a sequence of textual actions and observations.  The benchmark contains six task categories: Pick \& Place, Examine in Light, Clean \& Place, Heat \& Place, Cool \& Place, and Pick Two \& Place.  These tasks are well suited for evaluating credit assignment because successful completion often depends on a small number of critical actions mixed with many preparatory or redundant steps.

\subsection{WebShop}
WebShop evaluates web-based shopping agents.  The agent receives a user instruction and interacts with a simulated shopping website to search for products, inspect items, choose options, and make a final purchase.  We report both task success and task score.  Task success measures whether the agent completes a purchase, while task score evaluates how well the purchased item matches the instruction-specified requirements such as product attributes, options, type, and price.

\subsection{Baselines}
The main paper compares EFCA with closed-source LLM baselines, prompting-based agent methods, trajectory-level RL methods, and stepwise policy optimization methods.  When results are reused from prior work, we mark them in the main results table.  EFCA is applied as a plug-in return refinement module on top of the stepwise optimization pipeline, so improvements should be interpreted as gains from better credit assignment rather than from a different base optimizer.

\section{Additional Analysis}

\subsection{Interpretation of Reweighting Ratio}
The mean reweighting ratio indicates how strongly EFCA adjusts the base step-level return during training.  A ratio near $1.0$ means that EFCA behaves similarly to the underlying optimizer.  A ratio above $1.0$ indicates that more actions are supported by positive feedback evidence, while a ratio below $1.0$ indicates that many steps are being suppressed due to invalid, ineffective, or stagnant behavior.  The observed increase in the mean ratio over training suggests that the policy gradually produces more actions that receive favorable process-level evidence.

\subsection{Interpretation of Return Variance}
The standard deviation of reweighted returns is expected to be larger than that of base returns when EFCA successfully differentiates between high- and low-contribution steps.  This increased variance should not be interpreted as instability by itself.  Instead, it reflects more discriminative step-level supervision: useful actions receive stronger learning signals, while ineffective actions receive weaker signals.  The clipping bounds $w_{\min}$ and $w_{\max}$ are used to prevent this differentiation from producing excessively large updates.

\subsection{Ablation Interpretation}
The ablation results support the complementarity of the two process signals.  Removing the short-term feedback signal weakens the model's ability to identify whether the current action has immediate local effect.  Removing the state-history signal weakens the model's ability to penalize repeated ineffective behavior and short-horizon stagnation.  Combining both signals gives EFCA a multi-timescale view of credit assignment: final rewards provide long-term outcome supervision, feedback credit provides short-term execution supervision, and state-history credit provides medium-term efficiency supervision.

\section{Limitations and Future Directions}

The main insight of EFCA is that long-horizon agent learning should not treat environment interaction as a source of sparse terminal rewards only.  Even when scalar rewards are delayed, the interaction process often contains structured feedback about action validity, local progress, and short-horizon stagnation.  Our current implementation instantiates this insight with lightweight pattern-based feedback sets, which makes the credit signal transparent and easy to plug into existing optimizers.  A natural future direction is to move from surface-level pattern matching to semantic feedback understanding, where the agent can recognize equivalent progress or failure signals across different environments, modalities, and feedback styles.

A second direction is to broaden multi-timescale credit assignment beyond the two process signals studied here.  EFCA combines long-term outcome rewards, short-term feedback, and medium-term state-history evidence, but real agentic tasks may contain richer temporal structure, such as subgoal completion, reversible recovery actions, delayed enabling actions, and exploration steps that are locally unproductive but globally necessary.  Future work could develop causal or hierarchical credit models that distinguish these cases and assign credit according to the role each action plays in the broader interaction trajectory.

Finally, EFCA highlights a coupling between credit assignment and exploration.  Environment-grounded credit can only refine supervision for evidence that appears in collected trajectories; it does not by itself guarantee that the agent will visit states with informative feedback.  This suggests an important research direction: designing exploration strategies, curricula, or environment interfaces that actively expose useful process signals.  In this view, feedback is not merely a byproduct of interaction, but a resource that can be elicited, structured, and used to train more robust long-horizon agents.

\end{document}

%% file: tables/main_table.tex
% Main results table skeleton aligned with the style of GiGPO (arXiv:2505.10978).
% Required packages in the main tex file:
% \usepackage{booktabs}
% \usepackage{multirow}
% \usepackage{xcolor,colortbl}
% \usepackage{arydshln}

\begin{tabular}{llccccccccc}
\toprule[1.1pt]
\multirow{2}{*}{Type} & \multirow{2}{*}{Method} & \multicolumn{7}{c}{\textsc{ALFWorld}} & \multicolumn{2}{c}{\textsc{WebShop}} \\
\cmidrule(lr){3-9} \cmidrule(lr){10-11}
& & Pick  & Look & Clean  & Heat & Cool & Pick2 & All & Task Scores & Task Success Rates \\
\midrule

\rowcolor{blue!10}
\multicolumn{11}{c}{\textit{Closed-Source Models}} \\
\midrule
Prompting  & GPT-4o$^{\dagger}$ & 75.3 & 60.8 & 31.2 & 56.7 & 21.6 & 49.8 & 48.0 & 31.8 & 23.7  \\
Prompting  & Gemini-2.5-Pro$^{\dagger}$ & 92.8 & 63.3 & 62.1 & 69.0 & 26.6 & 58.7 & 60.3 & 42.5 & 35.9 \\
\midrule

\rowcolor{green!10}
\multicolumn{11}{c}{\textit{Qwen2.5-1.5B-Instruct}} \\
\midrule
Prompting & Qwen2.5$^{\dagger}$ & 5.9 & 5.5 & 3.3 & 9.7 & 4.2 & 0.0 & 4.1 & 23.1 & 5.2 \\
Prompting & ReAct$^{\dagger}$ & 17.4 & 20.5 & 15.7 & 6.2 & 7.7 & 2.0 & 12.8 & 40.1 & 11.3 \\
Prompting & Reflexion$^{\dagger}$ & 35.3 & 22.2 & 21.7 & 13.6 & 19.4 & 3.7 & 21.8 & 55.8 & 21.9 \\
\cdashline{1-11}[4pt/1.2pt]
RL Training & PPO$^{\dagger}$ & 64.8{\std{3.5}} & 40.5{\std{6.9}} & 57.1{\std{4.9}}& 60.6{\std{6.6}} & 46.4{\std{4.0}} & 47.4{\std{1.9}}& 54.4{\std{3.1}}& 73.8{\std{3.0}} & 51.5{\std{2.9}} \\
RL Training & GRPO$^{\dagger}$ & 82.89{\std{3.62}} & 82.14{\std{6.37}} & 73.86{\std{6.84}} & 78.57{\std{0.00}} & 77.78{\std{4.54}} & 71.43{\std{3.89}} & 77.86{\std{1.33}} & 84.73{\std{0.49}} & 71.35{\std{2.05}} \\
RL Training & GiGPO$^{\dagger}$ & 98.81{\std{1.68}} & 95.16{\std{3.89}} & 81.46{\std{0.56}}& 78.57{\std{0.00}} & 94.44{\std{0.00}} & 93.65{\std{5.94}}& 90.88{\std{0.97}}& 87.94{\std{0.43}} & 73.83{\std{2.30}} \\
RL Training & GraphGPO$^{\dagger}$ & 95.15{\std{1.62}} & 100.0{\std{0.00}} & 85.26{\std{2.58}}& 85.71{\std{5.83}} & 96.30{\std{2.61}} & 93.65{\std{2.24}}& 92.71{\std{1.32}}& 89.29{\std{1.48}} & \textbf{78.65{\std{3.86}}} \\
RL Training & HGPO$^{\dagger}$ & 97.53{\std{0.77}} & 81.20{\std{2.05}} & 99.60{\std{0.69}} & 100.00{\std{0.00}} & 85.63{\std{2.47}} & 82.57{\std{3.58}} & 91.99{\std{1.17}} & 85.56{\std{2.86}} & 71.54{\std{4.00}} \\
RL Training & EFCA & 98.98{\std{1.14}} & 81.32{\std{3.20}} & 98.52{\std{0.94}} & 98.90{\std{0.99}} & 94.39{\std{4.40}} & 93.18{\std{3.34}} & \textbf{95.31{\std{1.09}}} & \textbf{89.81{\std{1.71}}} & 75.91{\std{2.33}} \\
\midrule

\rowcolor{yellow!40}
\multicolumn{11}{c}{\textit{Qwen2.5-7B-Instruct}} \\
\midrule
Prompting & Qwen2.5$^{\dagger}$ & 33.4 & 21.6 & 19.3 & 6.9 & 2.8 & 3.2 & 14.8 & 26.4 & 7.8 \\
Prompting & ReAct$^{\dagger}$ & 48.5 & 35.4 & 34.3 & 13.2 & 18.2 & 17.6 & 31.2 & 46.2 & 19.5 \\
Prompting & Reflexion$^{\dagger}$ & 62.0 & 41.6 & 44.9 & 30.9 & 36.3 & 23.8 & 42.7 & 58.1 & 28.8  \\
\cdashline{1-11}[4pt/1.2pt]
RL Training & PPO$^{\dagger}$ & 92.3{\std{4.0}} & 64.0{\std{8.4}} & 92.5{\std{2.4}}& 89.5{\std{7.0}} & 80.3{\std{2.0}} & 68.8{\std{8.3}}& 80.4{\std{2.7}}& 81.4{\std{3.1}} & 68.7{\std{5.1}} \\
RL Training & GRPO$^{\dagger}$ & 88.98\std{5.30} &	91.98\std{4.43} & 77.89\std{4.58} &	78.57\std{0.00}	& 90.74\std{5.24}	 & 71.43\std{3.89} &	83.33\std{2.05} & 84.31{\std{1.27}} & 75.00{\std{2.78}} \\
RL Training & GiGPO$^{\dagger}$ & 97.53\std{1.75} &	100.0\std{0.00} &	83.98\std{1.32} &	90.48\std{6.73} &	94.44\std{0.00} &	100.0\std{0.00} &	94.27\std{1.33} & 86.72{\std{1.44}} & 78.38{\std{1.94}} \\
RL Training & GraphGPO$^{\dagger}$ & 100.0\std{0.00} &	100.0\std{0.00} &	91.40\std{1.50} &	92.86\std{5.83} &	94.44\std{0.00} &	92.06\std{2.24} &	95.31\std{1.10} & 86.94{\std{0.68}} & \textbf{80.31{\std{1.33}}} \\
RL Training & HGPO$^{\ddagger}$ & 97.53{\std{0.77}} & 81.20{\std{2.05}} & 99.60{\std{0.69}} & 100.00{\std{0.00}} & 85.63{\std{2.47}} & 82.57{\std{3.58}} & 91.99{\std{1.17}} & 85.56{\std{2.86}} & 71.54{\std{4.00}} \\
RL Training & EFCA & 97.93{\std{0.37}} & 94.43{\std{3.02}} & 99.56{\std{0.76}} & 96.40{\std{0.22}} & 92.45{\std{1.52}} & 94.14{\std{2.75}} &\textbf{ 96.03{\std{0.41}}} & \textbf{89.06{\std{1.63}}} & 78.91{\std{2.06}} \\

\bottomrule[1.1pt]
\end{tabular}

%% file: figures/ablation.tex
% PGFPlots figure for ablation results. Requires:
% \usepackage{pgfplots}
% \pgfplotsset{compat=1.18}
% \usetikzlibrary{plotmarks}
\begin{tikzpicture}
  \pgfplotsset{
    ablation/.style={
      width=0.35\textwidth,
      height=0.245\textwidth,
      ymin=89,
      ymax=97,
      ytick={89,90,91,92,93,94,95,96},
      grid=major,
      grid style={draw=gray!22, line width=0.25pt},
      axis line style={black!80, line width=0.8pt},
      tick style={black!80, line width=0.6pt},
      tick label style={font=\scriptsize},
      label style={font=\footnotesize},
      title style={font=\bfseries\footnotesize, yshift=-0.6ex},
      xlabel style={font=\footnotesize, yshift=0.2ex},
      ylabel style={font=\footnotesize},
      line width=1.2pt,
      mark size=2.2pt,
      every axis plot/.append style={
        color=blue!70!black,
        mark=o,
        mark options={solid, fill=white, line width=0.9pt},
      },
      clip=false,
    },
    ablation fill/.style={
      draw=none,
      fill=blue!13,
      forget plot,
    },
    bestmark/.style={
      only marks,
      mark=*,
      mark size=2.8pt,
      color=red!70!black,
      mark options={fill=red!70!black, draw=white, line width=0.4pt},
      forget plot,
    },
  }
  
  \begin{axis}[
    ablation,
    name=plotlambda,
    title={(a) Reweighting $\lambda$},
    xlabel={Reweighting strength},
    ylabel={ALFWorld},
    xmin=-0.06, xmax=0.86,
    xtick={0,0.2,0.4,0.6,0.8},
    xticklabels={0,0.2,0.4,0.6,0.8},
  ]
  \addplot[ablation fill] coordinates {(0,89) (0,91.99) (0.2,92.51) (0.4,93.03) (0.6,92.58) (0.8,92.06) (0.8,89)} -- cycle;
  \addplot coordinates {(0,91.99) (0.2,92.51) (0.4,93.03) (0.6,92.58) (0.8,92.06)};
  \addplot[bestmark] coordinates {(0.4,93.03)};
  \node[font=\scriptsize, anchor=south] at (axis cs:0,92.05) {91.99};
  \node[font=\scriptsize, anchor=south] at (axis cs:0.2,92.57) {92.51};
  \node[font=\scriptsize, anchor=south] at (axis cs:0.4,93.09) {93.03};
  \node[font=\scriptsize, anchor=south] at (axis cs:0.6,92.64) {92.58};
  \node[font=\scriptsize, anchor=south] at (axis cs:0.8,92.12) {92.06};
  % \node[font=\scriptsize\bfseries, red!70!black, fill=white, inner sep=1pt, anchor=south west] at (axis cs:0.43,93.62) {Best};
  \end{axis}
  
  \begin{axis}[
    ablation,
    at={(plotlambda.east)},
    xshift=0.06\textwidth,
    anchor=west,
    name=plotstate,
    title={(b) State-history $\beta$},
    xlabel={State-history coefficient},
    yticklabel=\empty,
    xmin=-0.12, xmax=1.72,
    xtick={0,0.4,0.8,1.2,1.6},
    xticklabels={0,0.4,0.8,1.2,1.6},
  ]
  \addplot[ablation fill] coordinates {(0,89) (0,90.95) (0.4,91.67) (0.8,95.31) (1.2,93.42) (1.6,89.91) (1.6,89)} -- cycle;
  \addplot coordinates {(0,90.95) (0.4,91.67) (0.8,95.31) (1.2,93.42) (1.6,89.91)};
  \addplot[bestmark] coordinates {(0.8,95.31)};
  \node[font=\scriptsize, anchor=south] at (axis cs:0,91.01) {90.95};
  \node[font=\scriptsize, anchor=south] at (axis cs:0.4,91.73) {91.67};
  \node[font=\scriptsize, anchor=south] at (axis cs:0.74,95.37) {95.31};
  \node[font=\scriptsize, anchor=south] at (axis cs:1.2,93.48) {93.42};
  \node[font=\scriptsize, anchor=south east] at (axis cs:1.58,89.97) {89.91};
  % \node[font=\scriptsize\bfseries, red!70!black, fill=white, inner sep=1pt, anchor=south west] at (axis cs:0.86,95.70) {Best};
  \end{axis}
  
  \begin{axis}[
    ablation,
    at={(plotstate.east)},
    xshift=0.06\textwidth,
    anchor=west,
    title={(c) Feedback $\alpha$},
    xlabel={Feedback coefficient},
    yticklabel=\empty,
    xmin=-0.15, xmax=2.15,
    xtick={0,0.5,1.0,1.5,2.0},
    xticklabels={0,0.5,1,1.5,2},
  ]
  \addplot[ablation fill] coordinates {(0,89) (0,91.40) (0.5,94.86) (1.0,93.03) (1.5,95.31) (2.0,94.27) (2.0,89)} -- cycle;
  \addplot coordinates {(0,91.40) (0.5,94.86) (1.0,93.03) (1.5,95.31) (2.0,94.27)};
  \addplot[bestmark] coordinates {(1.5,95.31)};
  \node[font=\scriptsize, anchor=south] at (axis cs:0,91.46) {91.4};
  \node[font=\scriptsize, anchor=south] at (axis cs:0.5,94.92) {94.86};
  \node[font=\scriptsize, anchor=south] at (axis cs:1.0,93.09) {93.03};
  \node[font=\scriptsize, anchor=south] at (axis cs:1.42,95.37) {95.31};
  \node[font=\scriptsize, anchor=south east] at (axis cs:1.98,94.33) {94.27};
  % \node[font=\scriptsize\bfseries, red!70!black, fill=white, inner sep=1pt, anchor=south west] at (axis cs:1.58,95.70) {Best};
  \end{axis}
  \end{tikzpicture}
  

%% file: figures/debug_step_reweight_ratio_mean_pgfplots.tex
% Single-column PGFPlots line chart for debug_step_reweight_ratio_mean.
% Requires in the main tex preamble:
% \usepackage{pgfplots}
% \pgfplotsset{compat=1.18}
% \usepgfplotslibrary{fillbetween}
\begin{tikzpicture}
\begin{axis}[
    width=0.7\columnwidth,
    height=0.38\columnwidth,
    xmin=1, xmax=250,
    ymin=0.65, ymax=1.60,
    xlabel={Training Step},
    ylabel={Reweight Ratio Mean},
    xtick={1,50,100,150,200,250},
    ytick={0.7,0.9,1.1,1.3,1.5},
    grid=major,
    grid style={draw=gray!22, line width=0.25pt, dotted},
    axis line style={black!80, line width=0.75pt},
    tick style={black!80, line width=0.55pt},
    tick label style={font=\small},
    label style={font=\small},
    legend style={
        draw=black!20,
        fill=white,
        fill opacity=0.85,
        text opacity=1,
        font=\footnotesize,
        at={(0.53,0.37)},
        anchor=north west,
    },
]
% Raw-value variation is shown as a local translucent band around the smoothed curve.
\addplot[name path=upper, draw=none, forget plot]
    table[x=Step, y=Upper] {figures/debug_step_reweight_ratio_mean_band.dat};
\addplot[name path=lower, draw=none, forget plot]
    table[x=Step, y=Lower] {figures/debug_step_reweight_ratio_mean_band.dat};
\addplot[blue!18, opacity=0.70, forget plot]
    fill between[of=upper and lower];

% Smoothed curve is used as the main solid line.
\addplot[
    blue!70!black,
    line width=1.25pt,
    mark=none,
] table[x=Step, y=Ratio] {figures/debug_step_reweight_ratio_mean_smooth.dat};
\addlegendentry{Smoothed ratio}

\addplot[
    black!55,
    dashed,
    line width=0.7pt,
    mark=none,
] coordinates {(1,1.0) (250,1.0)};
\addlegendentry{No reweighting}
\end{axis}
\end{tikzpicture}

%% file: figures/debug_base_step_return_std_pgfplots.tex
% Single-column PGFPlots line chart for base and reweighted step return standard deviation.
% Requires in the main tex preamble:
% \usepackage{pgfplots}
% \pgfplotsset{compat=1.18}
% \usepgfplotslibrary{fillbetween}
\begin{tikzpicture}
\begin{axis}[
    width=0.72\columnwidth,
    height=0.40\columnwidth,
    xmin=1, xmax=250,
    ymin=0.4, ymax=6.0,
    xlabel={Training Step},
    ylabel={Step Return Std.},
    xtick={1,50,100,150,200,250},
    ytick={1,2,3,4,5,6},
    grid=major,
    grid style={draw=gray!22, line width=0.25pt, dotted},
    axis line style={black!80, line width=0.75pt},
    tick style={black!80, line width=0.55pt},
    tick label style={font=\small},
    label style={font=\small},
    legend style={
        draw=black!20,
        fill=white,
        fill opacity=0.85,
        text opacity=1,
        font=\footnotesize,
        at={(0.53,0.27)},
        anchor=north west,
    },
]
% Raw-value variation is shown as translucent local bands around each smoothed curve.
\addplot[name path=baseupper, draw=none, forget plot]
    table[x=Step, y=BaseUpper] {figures/debug_base_step_return_std_band.dat};
\addplot[name path=baselower, draw=none, forget plot]
    table[x=Step, y=BaseLower] {figures/debug_base_step_return_std_band.dat};
\addplot[blue!16, opacity=0.68, forget plot]
    fill between[of=baseupper and baselower];

\addplot[name path=rewupper, draw=none, forget plot]
    table[x=Step, y=ReweightedUpper] {figures/debug_base_step_return_std_band.dat};
\addplot[name path=rewlower, draw=none, forget plot]
    table[x=Step, y=ReweightedLower] {figures/debug_base_step_return_std_band.dat};
\addplot[red!14, opacity=0.62, forget plot]
    fill between[of=rewupper and rewlower];

% Smoothed curves are used as the main solid lines.
\addplot[
    blue!70!black,
    line width=1.2pt,
    mark=none,
] table[x=Step, y=BaseStd] {figures/debug_base_step_return_std_smooth.dat};
\addlegendentry{Base step return}

\addplot[
    red!70!black,
    line width=1.2pt,
    mark=none,
] table[x=Step, y=ReweightedStd] {figures/debug_base_step_return_std_smooth.dat};
\addlegendentry{Reweighted step return}
\end{axis}
\end{tikzpicture}

%% file: figures/computational_cost_pgfplots.tex
\begin{tikzpicture}
\begin{axis}[
    width=0.95\linewidth,
    height=0.58\linewidth,
    xlabel={Training Step},
    ylabel={Time per Step (s)},
    xmin=1, xmax=250,
    ymin=150, ymax=720,
    grid=both,
    grid style={line width=.1pt, draw=gray!20},
    major grid style={line width=.2pt,draw=gray!35},
    legend style={at={(0.98,0.98)},anchor=north east,font=\scriptsize,draw=none,fill=white,fill opacity=0.85,text opacity=1},
    tick label style={font=\scriptsize},
    label style={font=\small},
]
\addplot+[blue!25, line width=0.35pt, mark=none] table[x=Step,y=EFCA,col sep=comma] {figures/computational_cost_timing.csv};
\addlegendentry{EFCA raw}
\addplot+[blue!85!black, line width=1.25pt, mark=none] table[x=Step,y=EFCA_smoothed,col sep=comma] {figures/computational_cost_timing.csv};
\addlegendentry{EFCA smoothed}
\addplot+[red!25, line width=0.35pt, mark=none] table[x=Step,y=HGPO,col sep=comma] {figures/computational_cost_timing.csv};
\addlegendentry{HGPO raw}
\addplot+[red!85!black, line width=1.25pt, mark=none] table[x=Step,y=HGPO_smoothed,col sep=comma] {figures/computational_cost_timing.csv};
\addlegendentry{HGPO smoothed}
\end{axis}
\end{tikzpicture}